\documentclass[runningheads]{llncs}

\usepackage{microtype}
\usepackage{graphicx}
\usepackage{booktabs} 
\usepackage{amsmath, amssymb}
\usepackage{mathtools}
\usepackage{subcaption}
\usepackage{xcolor}
\usepackage{thm-restate}
\usepackage{float}
\usepackage[misc]{ifsym}

\usepackage{hyperref}

\newcommand{\Xcomment}[1]{{}}
\newcommand{\eps}{\epsilon}

\newcommand{\noaccents}[1]{#1}
\newcommand{\newagentvar}[3][\noaccents]{%
\expandafter\newcommand\expandafter{\csname #2\endcsname}{#1{#3}}%
\expandafter\newcommand\expandafter{\csname #2s\endcsname}{#1{\boldsymbol{#3}}}%
\expandafter\newcommand\expandafter{\csname #2smi\endcsname}[1][i]{#1{\boldsymbol{#3}}_{-##1}}%
\expandafter\newcommand\expandafter{\csname #2i\endcsname}[1][i]{#1{#3}_{##1}}%
\expandafter\newcommand\expandafter{\csname #2ith\endcsname}[1][i]{#1{#3}_{(##1)}}%
}

\newcommand{\newvecagentvar}[3][\noaccents]{%
\expandafter\newcommand\expandafter{\csname #2\endcsname}{#1{#3}}%
\expandafter\newcommand\expandafter{\csname #2s\endcsname}{#1{\boldsymbol{#3}}}%
\expandafter\newcommand\expandafter{\csname #2smi\endcsname}[1][i]{#1{\boldsymbol{#3}}_{-##1}}%
\expandafter\newcommand\expandafter{\csname #2i\endcsname}[1][i]{#1{\boldsymbol{#3}}_{##1}}%
\expandafter\newcommand\expandafter{\csname #2ith\endcsname}[1][i]{#1{#3}_{(##1)}}%
}

\newvecagentvar{strat}{x}
\newvecagentvar[\bar]{avgstrat}{\strat}
\newagentvar{payoff}{p}
\newagentvar{augpayoff}{\hat{p}}
\newvecagentvar{supergrad}{q}

\DeclarePairedDelimiterX{\inp}[2]{\langle}{\rangle}{#1, #2}
\newvecagentvar{point}{z}
\newvecagentvar{dynamic}{g}
\newvecagentvar{operator}{v}
\newvecagentvar{secondstrat}{w}

\newcommand{\norm}[1]{\|{#1}\|}

\begin{document}

\title{Exponential Convergence of Gradient Methods in Concave Network Zero-sum Games \thanks{The research was funded by an NSERC Discovery Grant, an NSERC Discovery Acceleration Grant, and Canadian Research Chair stipend.}}
\titlerunning{Convergence in Concave Network Zero-sum Games}
%
\author{Amit Kadan\inst{1} \and
Hu Fu\inst{1}[\Letter]}
\tocauthor{Amit Kadan and Hu Fu}
\toctitle{Exponential Convergence of Gradient Methods in Concave Network Zero-sum Games}
\authorrunning{A. Kadan and H. Fu}
%
\institute{University of British Columbia, Vancouver BC V6T 1Z4, Canada \\ \email{\{amitkad, hufu\}@cs.ubc.ca}}
\maketitle              
\begin{abstract}
    Motivated by Generative Adverserial Networks, we study the computation of Nash equilibrium in concave \emph{network zero-sum games} (NZSGs), a multiplayer generalization of two-player zero-sum games first proposed with linear payoffs. Extending previous results, we show that various game theoretic properties of convex-concave two-player zero-sum games are preserved in this generalization. We then generalize last iterate convergence results obtained previously in two-player zero-sum games. We analyze convergence rates when players update their strategies using Gradient Ascent, and its variant, Optimistic Gradient Ascent, showing last iterate convergence in three settings --- when the payoffs of players are linear, strongly concave and Lipschitz, and strongly concave and smooth. We provide experimental results that support these theoretical findings.

\keywords{Network zero-sum games \and Last iterate convergence \and Convergence of gradient ascent \and Generative adversarial networks.}
\end{abstract}

\section{Introduction}
\label{sec:intro}
Connections between game theory and learning had long been known, before interest resurged recently in the machine learning community, largely due to the success of Generative Adversarial Networks (GANs), a novel framework for learning generative models~\cite{goodfellow2014generative}. 
A GAN is formulated as a two-player zero-sum game between two neural networks, a generator and a discriminator. 
The generator attempts to fool the discriminator by mapping random noise to images that look similar to samples from a target distribution, while the discriminator learns to distinguish the generator's output from real samples from the target distribution. 
Theoretically, at the equilibrium of this game, the generator outputs the target distribution.
In practice, GANs produce promising results on a number of tasks including image generation, semantic segmentation, and text-to-image synthesis \cite{goodfellow2016nips}.

 Among many theoretical questions opened up by GANs, that of \emph{last iterate} convergence has attracted much attention and seen exciting progress.
 Classical results show that, when players use \emph{no-regret online learning} algorithms to play a two-player zero-sum game, 
 the \emph{time average} of their strategies converge to a Nash equilibrium --- a point where neither player can make gains by unilaterally deviating from their current strategy.  
In GANs, strategies correspond to parameters of neural networks; 
averaging strategies makes little sense.
It is therefore desirable that the players' strategies, from iteration to iteration, should converge to an equilibrium.  
This is known as \emph{last iterate} convergence, which is not implied by classical results.
A number of simple algorithms have been shown to give rise to such convergence in two-player zero-sum games, with exponential convergence rates in various settings \cite{gidel2018variational,liang2018interaction,mokhtari2019unified} (see Section~\ref{sec:related} for more details).

 On the other hand, many recently proposed extensions of GANs go beyond the two-player zero-sum framework, either to address challenges faced by the original GAN, or to make it more versatile. 
 In particular, many models introduce more agents (neural networks) to the game. 
 For example, Hoang et al.\@ \cite{hoang2018mgan} proposed using an ensemble of generators to address \emph{mode collapse}, a common problem of the classical GAN, where the generator captures only one or few modes of the data distribution.
 Other architectures incorporate a third classifying network, which is in direct competition with either the generator \cite{vandenhende2019three} or the discriminator \cite{chongxuan2017triple}; 
 such architectures are often built for semi-supervised learning.
 Lastly, some architectures incorporate an additional encoding network which, like the generator, competes with the discriminator, and allows for sampling from a latent distribution that encodes additional information about the data distribution \cite{brock2016neural,che2016mode,donahue2016adversarial}.
Results on two-player zero-sum games do not apply to these architectures with more than two agents.
It is also well known that two-player zero-sum games have many properties not extensible to games with more players or non zero-sum payoffs.

We observe that the extensions above all give rise to \emph{network zero-sum games} (NZSGs), a class of games first proposed and studied by Cai et al.~\cite{cai2016zero}.  
An NZSG is structured by a graph, where each node corresponds to a player, and along each edge a game is played between its two node players. 
A player chooses one strategy to be used in all the games in which which she is engaged; the sum of all players' payoffs is always zero.  
Since players cannot choose different strategies for different games, an NZSG is not a simple parallelization of multiple two-player zero-sum games.
However, Cai et al.~\cite{cai2016zero} showed that NZSGs with linear payoffs preserve certain properties from two-player zero-sum games. 
In particular a Nash in an NZSG can be computed via a linear program. 

We first generalize results of \cite{cai2016zero} on the tractability of equilibrium for NZSGs (Section~\ref{sec:comp-nash}); we show that in an NZSG with concave payoffs, a Nash can be computed via no-regret learning.
Then, as our main result, we show last iterate convergence results for NZSGs with several classes of payoffs (Section~\ref{sec:last-iterate}), when players adopt simple learning rules used in practice, such as \emph{Gradient Ascent} (GA) and \emph{Optimistic Gradient Ascent} (OGA).  
GA is the most ubiquitous optimization algorithm.  
It may be seen as a smoothed best response, and so it may not be surprising that it produces dynamics that diverge from the equilibrium in two-player zero-sum games with linear payoffs~\cite{daskalakis2018limit,liang2018interaction}.  
We show that this phenomenon persists in NZSGs with linear payoffs. 
OGA, on the other hand, incorporates some minimal memory, and uses information from one step before.  
This small tweak has been shown to induce last iterate convergence in two-player zero-sum games with either linear payoffs or strongly concave payoffs that are smooth in various senses~\cite{daskalakis2017training,liang2018interaction,mokhtari2019unified}.
We extend these to NZSGs, showing comparable convergence performance.  
For two-player zero-sum games with strongly concave payoffs, GA is known to induce last iterate convergence; we generalize this as well.

We use two sets of tools.  
Our main tool for NZSGs with linear payoffs is dynamical systems. 
Strategies played in a repeated game give rise to a dynamical system; techniques for analyzing such systems naturally can be used to analyze various update algorithms~\cite{daskalakis_et_al:LIPIcs:2018:10120,daskalakis2018limit,liang2018interaction}.
Our results on both the divergence of GA and convergence of OGA dynamics are built on linear algebraic techniques used to analyze the corresponding dynamical systems. 
Crucial to the arguments is an algebraic property we show for NZSGs; namely, that a Hessian matrix associated with the payoff functions is antisymmetric everywhere.

We use Lyapunov-style convergence proofs to show results in NZSGs with strongly concave and smooth payoffs.
Apart from existing arguments for two-player zero-sum games, our proof exploits a structural lemma (Lemma~\ref{lemma:util-sum}), which may be of independent interest.

In Section~\ref{sec:exp}, we provide experiments that validate our theoretical findings.

\subsection{Related Work}\label{sec:related}
Cai et al.~\cite{cai2016zero} introduced the class of network zero-sum games, and showed that a Nash equilibrium of an NZSG can be computed by a linear programming when each player's strategy is a distribution over a finite number of actions.

A few papers study convergence in $n$-player games. 
The most closely related work to ours is Azizian et al.~\cite{Azizian2020smooth}. 
They show that various gradient-based algorithms, including \ref{alg:oga}, converge at an exponential rate to the Nash in a class of smooth and ``monotone'' $n$-player games. 
With slight modification explained in the technical sections, our results on the OGA dynamics in NZSGs with strongly concave and smooth payoffs or with linear payoffs can be obtained by showing these games to be smooth and monotone.  
Our proofs in these settings may be viewed as alternative approaches to showing these results.
An advantage of our approach is that it is readily modified to apply for games with Lipschitz payoffs, as we demonstrate in Section~\ref{subsec:strongc+lip}.
 
Balduzzi et al.~\cite{BalduzziRMFTG18} study two classes of $n$-player games, Hamiltonian games and potential games, both of which are specific instances of NZSGs.  
They show that, when players use a continuous-time version of GA to update their strategies in a Hamiltonian game, the dynamics circle perpetually around the Nash of the game. 
They propose \emph{Symplectic Gradient Adjustment} (SGA) and show it to converge in last iterate for both Hamiltonian and potential games. 
Balduzzi et al.~\cite{Balduzzi2020Smooth} study another class of games called Smooth Market Games, which consist of payoffs that are pairwise zero-sum. 
They show that a continuous time version of GA converges in last iterate to the Nash of a game when payoffs are strictly concave in players' strategies.

A number of papers study last iterate convergence in concave two-player zero-sum games. 
Liang and Stokes~\cite{liang2018interaction} use tools from dynamical systems to show exponential convergence of the last iterate in bilinear games when players use \ref{alg:oga}. 
They also show exponential convergence of the last iterate in games with smooth and strongly concave payoffs when players use \ref{alg:ga}. 
Mokhtari et al.~\cite{mokhtari2019unified} show exponential convergence of the last iterate in games with bilinear, or smooth and strongly concave payoffs when the players use \ref{alg:oga}, by interpreting \ref{alg:oga} as an approximation of the \emph{Proximal Point Method}.
Gidel et al.~\cite{gidel2018variational} use a variational inequality perspective to show exponential convergence of a variant of OGA in constrained two-player zero-sum games with smooth and strongly concave payoffs.
Merkitopolous et al.~\cite{mertikopoulos2018optimistic} use similar tools to show last iterate, but not exponential convergence for Mirror Descent and Optimistic Mirror Descent
when payoffs are strongly concave, and for Optimistic Mirror Descent when payoffs are linear.

\subsection{Notations and Mathematical Conventions}

Vectors in $\mathbb{R}^k$ are denoted by boldface, and scalars by lowercase. 
Time indices are denoted by superscripts, while players are identified by subscripts. 
For a square matrix $A$ we denote the set of its eigenvalues by $\lambda(A)$.  
$I_m$ denotes the $m \times m$ identity matrix.

\begin{definition}
Given $U \subset \mathbb{R}^k$, and concave function $f: U \to \mathbb{R}$. $\mathbf{q} \in \mathbb{R}^k$ is a \emph{supergradient} of $f$ at $\mathbf{u}$ if $\forall \mathbf{u}' \in U$,
    $f(\mathbf{u}') \leq f(\mathbf{u}) + \inp{\mathbf{q}}{\mathbf{u}' - \mathbf{u}}$.
The set of supergradients of $f$ at a point $\mathbf{u}$ is denoted by $\partial f (\mathbf{u})$.
\end{definition}

\begin{definition}
For $\alpha > 0$, a function $f: U \to \mathbb{R}$ is $\alpha$-strongly concave if $\forall \mathbf{u}, \mathbf{u}' \in U$ and $\mathbf{q} \in \partial f(\mathbf{u})$,
\begin{align*}
    &f(\mathbf{u}') \leq f(\mathbf{u}) + \inp{\mathbf{q}}{\mathbf{u}' - \mathbf{u}} - \frac{\alpha}{2}\norm{\mathbf{u} - \mathbf{u}'}^2.
\end{align*}
A function $g: U \times V \to \mathbb R$ is $\alpha$-strongly concave in~$\mathbf{u}$ if for any $\mathbf{v} \in V$, $h(\mathbf{u}) \coloneqq g(\mathbf{u}, \mathbf{v})$ is $\alpha$-strongly concave.
\end{definition}

\section{Network Zero-sum Games Basics}\label{sec:comp-nash}

In this section we extend network zero-sum games as defined by  Cai et al.~\cite{cai2016zero} to allow continuous action spaces.
We then show that in games with concave payoff functions, an equilibrium can be efficiently computed with no-regret dynamics. 

\begin{definition}\label{def:nzsg} A network game $\mathcal{G}$ consists of the following:
\begin{itemize}
    \item a finite set $V=\{1, ..., n\}$ of players, and a set $E$ of edges which are unordered pairs of players $[i, j], i \neq j$;
    \item for each player $i \in V$, a convex set $X_i \subseteq \mathbb{R}^{d_i}$, the strategy set for player $i$;
    \item for each edge $[i,j] \in E$, a two-person game $(\payoff_{ij}, \payoff_{ji})$, where $\payoff_{ij}: X_i \times X_j \to \mathbb{R}$, and $p_{ji}: X_j \times X_i \to \mathbb R$.
\end{itemize}

    Given a strategy profile $\strats = (\strati[1], ..., \strati[n]) \in \prod_{j \in V}X_j$, player $i$'s \emph{payoff} is $\payoffi(\strats) \coloneqq \sum_{[i,j] \in E}\payoff_{ij}(\strati, \strati[j])$.
    
 A network game is a \emph{network zero-sum game} (NZSG) if for all strategy profiles $\strats \in \prod_{j \in V}X_j$,
$\sum_{i \in V}\payoffi(\strats) = 0$. 
\end{definition}

We let $X$ denote $\prod_{i \in V} X_i$, and $d \coloneqq \sum_{i \in V}{d_i}$.
\emph{Two-player zero-sum games} are special cases of NZSGs, where $V$ has two nodes, connected by one edge.

In a \emph{concave NZSG}, each $\payoffi[ij](\strati, \strati[j])$ is concave in $\strati$. 
An NZSG is \emph{linear} if each $\payoffi[ij](\strati, \strati[j])$ is linear in both $\strati$ and~$\strati[j]$.

Let $\stratsmi$ denote the strategy profile without player $i$'s strategy, i.e., $\stratsmi = (\strati[1], ..., \strati[i-1], \strati[i+1], ..., \strati[n])$.

\begin{definition}\label{def:nash}
A strategy profile $\strats^*$ is a \emph{Nash equillibrium} for an NZSG if for each player~$i$,  for any strategy $\strati \in X_i$, 
$\payoffi(\strats^*) \geq  \payoffi(\strati, \stratsmi^*)$. 
\end{definition}

It can be shown via a fixed point argument that, in a concave NZSG where each player's strategy space $X_i$ is convex and compact, a Nash equilibrium always exists~\cite{menache2011network}.
Cai et al.~\cite{cai2016zero} showed that for linear NZSGs where each player's strategy set is a simplex, a Nash can be computed efficiently by a linear program.

As a warm-up, we show that another classical technique for computing equilibrium in two-player zero-sum games, namely, no-regret learning algorithms, can be used to find an approximate Nash in general concave NZSGs.

Given an NZSG with compact strategy sets, consider the players playing it repeatedly.  
Let $\strati^s$ and $\stratsmi^s$ denote, respectively, player~$i$'s and the other players' strategies at time step~$s$ of the game.  
Each player should only respond to the past strategies of her opponents; i.e., $\strati^s$ may depend only on $\stratsmi^1, \ldots, \stratsmi^{s-1}$.

\begin{definition}\label{def:no-regret}
In a repeated game, a player's \emph{regret} at time $t$, $r_i(t)$, is 
\begin{align*}
    r_i(t) = \max_{\strati \in X_i}\sum_{s=1}^t \left[\payoffi(\strati, \stratsmi^s) - \payoffi(\strati^s, \stratsmi^s) \right].
\end{align*}
A player~$i$'s strategy $(\strati^s)_s$ 
has \emph{no-regret} if for all~$t$,  
   $\frac{1}{t}r_i(t) \leq \epsilon(t)$ for some $\epsilon(t) \to 0$ as $t \to \infty$.
  An algorithm that produces no-regret strategies is a no-regret algorithm. 
\end{definition}

It is well known that efficient no-regret algorithms exist~\cite{cesa2006prediction}, and that in a two-player zero-sum game, if players use no-regret dynamics, the \emph{time average} of their strategies converges to a Nash equilibrium~\cite{cesa2006prediction}.  
We show this phenomenon generalizes to NZSGs with concave payoffs.

\begin{restatable}{proposition}{propnash}\label{prop:nash} In a concave NZSG with compact strategy sets, if each player uses strategies that have no-regret, then the strategy profile 
where each player plays her time-average strategy 
converges to a Nash equilibrium.
\end{restatable}


A key step in the proof of Proposition~\ref{prop:nash} is the following property of NZSGs.  
We will make repeated use of this property later in the paper.

\begin{restatable}{lemma}{lemutilsum}
\label{lemma:util-sum}
In an NZSG, for any two strategy profiles $\strats$ and~$\strats^*$, we have
\begin{align*}
\sum_i \payoffi(\strati, \stratsmi^*) = - \sum_i \payoffi(\strati^*, \stratsmi).
\end{align*}
\end{restatable}



As we discussed in the Introduction, Proposition~\ref{prop:nash} is not adequate for applications where strategies are parameters of neural networks, since taking 
 averages over strategies makes little sense in  such settings.
Following much recent literature, we shift the focus to last iterate convergence.


\section{Last Iterate Convergence in NZSGs}
\label{sec:last-iterate}
In this section we present our main results on last-iterate convergence in NZSGs when players use gradient style updates. 
In this section we assume that the strategy spaces are unconstrained, i.e., $X_i = \mathbb{R}^{d_i}$ for each~$i$.

We first formally define the two update rules we focus on. 
Recall that we use $\strati^t$ to denote player~$i$'s strategy at time~$t$.
A player using \emph{Gradient Ascent} (GA) modifies her strategy by
\begin{align*}\label{alg:ga}
    \strati^{t+1} = \strati^t + \eta\nabla_{\strati}\payoffi(\strats^t),\tag{GA}
\end{align*}
where $\eta > 0$ is a fixed step size.  
A player using \emph{Optimistic Gradient Ascent} (OGA) updates her strategy by
\begin{align*}
\label{alg:oga}
    \strati^{t+1} = \strati^t + 2\eta\nabla_{\strati}\payoffi(\strats^t) - \eta\nabla_{\strati}\payoffi(\strats^{t-1}),
    \tag{OGA}
\end{align*}
where $\eta > 0$ again is a fixed step size.

\subsection{Linear NZSGs}\label{subsec:linear}
Even in a two-player zero-sum bilinear game, 
i.e., $\payoffi[1](\strati[1], \strati[2]) = - \payoffi[2] (\strati[2], \strati[1]) = \payoff(\strati[1], \strati[2]) = \strati[1]^{\top}C\strati[2]$,  where $C$ is a $d_1 \times d_2$ matrix, 
if each player uses \ref{alg:ga}, over time the players' strategies diverge from the set of Nash~\cite{liang2018interaction}. 
If, instead, players use OGA,
their strategies converge to a Nash of the game~\cite{daskalakis2017training,liang2018interaction,mertikopoulos2018optimistic}.
We show that these phenomena continue to hold for linear NZSGs.

To state the rates of convergence and divergence, we need to introduce a matrix~$H$ for a linear NZSG, which we motivate later.
Given an NZSG and a strategy profile~$\strats$, the \emph{Hessian} $H(\strats)$ is a $d \times d$ block matrix with the $(i,j)^{th}$ block given by 
\begin{align*}
    H_{ij}(\strats) = \nabla^2_{\strati[j], \strati}\payoffi(\strats).
\end{align*}

Denote the smallest nonzero modulus of an eigenvalue of $H$ by $\omega(H)$, and denote the largest modulus of an eigenvalue of~$H$ by $\rho(H)$. Denote the distance to a set by $d(\mathbf{u}, S) \coloneqq \min_{\mathbf{s} \in S}\norm{\mathbf{u} - \mathbf{s}}$. 

\begin{restatable}{theorem}{proplinga}\label{prop:lin-gd}
Consider an unconstrained, linear NZSG. Let $X^*$ denote the set of Nash of the game.
Assume each player uses \ref{alg:ga} to update her strategy at each time step. 
Assume $d(\strats^0, X^*) \geq R$, for some $R > 0$. Then at each time step~$t$,
\begin{align*}
    d(\strats^{t}, X^*)^2 \geq  (1 + \eta^2\omega(H)^2)^{t} R^2.
\end{align*}
\end{restatable}


\begin{restatable}{theorem}{proplinoga}\label{prop:lin-ogd}
Consider an unconstrained, linear NZSG. Assume that each player uses \ref{alg:oga} as her update rule. Let $X^*$ denote the set of Nash of the game.
If $H(\strats)$ is diagonalizable for all $\strats$, and if $d(\strats^0, X^*) \leq r, d(\strats^1, X^*) \leq r$ for some $r > 0$.
Then setting $\eta = 1 / 2 \rho(H)$, at each time step $t$,
\begin{align*}
    d(\strats^{t+1}, X^*)^2 \leq \left(\frac{1}{2} + \frac{1}{2}\left(1 - \left( \frac{\omega(H)}{\rho(H)}\right)^2 \right)^{\frac{1}{2}}\right)^t r^2.
\end{align*}
\end{restatable}


We sketch the proof ideas and relegate details 
to the supplementary file.
We formulate the behavior of GA and OGA as trajectories of \emph{dynamical systems}; 
this view has been taken in several previous works, which also analyze the behaviors of updating algorithms using tools from dynamical systems 
\cite{daskalakis_et_al:LIPIcs:2018:10120,daskalakis2018limit,liang2018interaction}.

\begin{definition}
A relation of the form $\strats^{t+1} = \dynamics(\strats^t)$, also written as $\strats \mapsto \dynamics(\strats)$, is a discrete time dynamical system with update rule $\dynamics: \mathbb R^k \to \mathbb R^k$.
A point $\points$ is a \emph{fixed point} of $\dynamics$ if $\dynamics(\points) = \points$.
\end{definition}

If players use \ref{alg:ga}, the strategies evolve according to the dynamical system  $\strats^{t+1} = (I_d + \eta H)\strats^t$, where $H$ is the Hessian matrix defined above. 
It is not hard to show that the set of Nash equilibria is precisely the set of fixed points of this dynamical system.  
Note that, when $\dynamics$ is a linear function, as is the case for the GA dynamics, a point is its fixed point if and only if it is in the eigenspace of~$\dynamics$ for eigenvalue~$1$. 

For a dynamical system with update rule $\dynamics: \mathbb{R}^k \to \mathbb{R}^k$, the \emph{Jacobian} is the matrix with its $(i,j)^{\text{-th}}$ entry $J_{ij} = \frac{\partial \dynamici}{\partial \strats_j}$.
The eigenvalues of the Jacobian~$J$ at a fixed point~$\points$ describe the behavior of the dynamics around~$\points$ .  
Roughly speaking, if all eigenvalues of~$J$ have modulus greater than~$1$, then in a neighborhood around~$\points$, the dynamics diverges from~$\points$; conversely, if all eigenvalues of~$J$ have modulus smaller than~$1$, in a neighborhood of~$\points$ the dynamics converges to~$\points$. 
When $\dynamics$ is linear, this characterization of convergence/divergence extends to the entire space (beyond neighborhoods around~$\points$), and allows some eigenvalues to be~$1$.


\begin{restatable}{proposition}{proplinconv}\label{prop:linear-convergence}
Let $Z$ denote the set of fixed points of a dynamical system with linear update rule: $\dynamics(\points) = J\points$, where J is diagonalizable.  
Let $d(\strats, Z)$ denote $\min_{\points \in Z} \norm{\strats- \points}$.
\begin{enumerate}
    \item[(a)] If $\forall \lambda \in \lambda(J)$ either $|\lambda| < 1$ or $\lambda = 1$, then letting $\sigma_{\max_{< 1}}(J)$ denote the largest modulus of any eigenvalue of $J$ not equal to 1, $\forall \strats^0 \in X$, $ d(\strats^{t+1}, Z) \leq (\sigma_{\max_{<1}}(J))^t d(\strats^0, Z)$.
    \item[(b)] If $\forall \lambda \in \lambda(J)$ either $|\lambda| > 1$ or $\lambda = 1$,  then letting $\sigma_{\min_{>1}}(J)$ denote the smallest modulus of any eigenvalue of $J$ not equal to 1, $\forall \strats^0 \in X, d(\strats^{t+1}, Z) \geq (\sigma_{\min_{>1}}(J))^t d(\strats^0, Z)$.
\end{enumerate}
\end{restatable}

To show Theorem~\ref{prop:lin-gd}, therefore, it suffices to analyze the eigenvalues of the matrix $J = I_d + \eta H$.  
The crucial observation is that, for NZSGs, the Hessian~$H$ is an antisymmetric matrix of the form 
\begin{align*}
 H = \begin{bmatrix} 0 &  C_{12} & \hdots & C_{1n} \\ -C_{12}^{\top} & 0 & \hdots &C_{2n} \\ \vdots & \vdots & \ddots & \vdots\\ -C_{1n}^{\top} & - C_{2n}^{\top} & \hdots & 0 \end{bmatrix}.
\end{align*}
This is a consequence of the following lemma on NZSGs in general:
\begin{restatable}{lemma}{lemmhesssym}
In an NZSG, if each $\payoffi$ has continuous second partial derivatives, then
\begin{align*}
     \nabla_{\strati,\strati[j]}^2\payoffi[ji](\strats) = -(\nabla_{\strati[j],\strati}^2\payoffi[ij](\strats))^{\top}.
\end{align*}
\end{restatable}
As a result, for the GA dynamics in a linear NZSG, all eigenvalues of~$H$ are imaginary, and therefore all the eigenvalues of $I_d + \eta H$ are of the form $1 + i \eta \lambda$ for some $\lambda \in \mathbb R$.  
Part~(b) of Proposition~\ref{prop:linear-convergence} indicates a diverging dynamics.

The antisymmetry of the Hessian~$H$ is also a crucial step in the proof of Theorem~\ref{prop:lin-ogd}. 
We first need to augment the state space to allow the memory from a previous step to be passed as part of the state.  
Following Daskalakis and Panageas~\cite{daskalakis2018limit}, we consider a dynamical system with the following update rule $\dynamics : \mathbb{R}^{2d} \to \mathbb{R}^{2d}$, defining  $\augpayoffi(\strats, \strats') \coloneqq \payoffi(\strats)$:
\begin{align}\label{eq:oga}
    &\dynamics(\strats, \strats') = (\dynamics^1(\strats, \strats'), \dynamics^2(\strats, \strats')), \\
    &\dynamici^1(\strats, \strats') = \strati + 2\eta\nabla_{\strati} \augpayoffi(\strats, \strats') - \eta\nabla_{\strati'} \augpayoffi(\strats, \strats'),  \nonumber\\
    &\dynamici^2(\strats, \strats') = \strati.  \nonumber 
\end{align}
More explicitly, for the \ref{alg:oga} update rule, we have the relation $(\strats^{t+1},\strats^t) = \dynamics(\strats^t, \strats^{t-1})$. 
We make use of a connection established by \cite{daskalakis2018limit} between the GA dynamics and the OGA dynamics (Proposition~\ref{prop:ga-to-oga}).
Besides another application of the antisymmetry of~$H$, we also use an expression for the determinant of a $2 \times 2$ block matrix (Lemma~\ref{lemma:schur}).

\begin{restatable}[\cite{daskalakis2018limit}]{proposition}{propgatooga}\label{prop:ga-to-oga}
Let $\points$ be a fixed point of the \ref{alg:ga} dynamics. 
Then, $(\points, \points)$ is a fixed point of the \ref{alg:oga} dynamics, and for each $\mu \in \lambda(J_{\mathrm{GA}})$ we have two eigenvalues in $\lambda(J_{\mathrm{OGA}})$ that are the roots of the quadratic equation
\begin{align*}
    \lambda^2 - (2\mu - 1)\lambda + (\mu -1) = 0.
\end{align*}
\end{restatable}
\begin{lemma}[\cite{gallier2010schur}]\label{lemma:schur}
Let $A$ be a block matrix of the following form
\begin{align*}
    A = \begin{bmatrix}
    M_1 & M_2 \\
    M_3 & M_4\end{bmatrix}, 
\end{align*}
where each $M_i$ is a square matrix, and $M_4$ is invertible. 
Then the determinant of $A$ is equal to the determinant of its Schur Complement:
\begin{align*}
    \det(A) = \det(M_1 - M_2(M_4)^{-1}M_3)\det(M_4).
\end{align*}
\end{lemma}

In order to be able to apply Proposition~\ref{prop:linear-convergence}, we make an additional diagonalizability assumption on $H$. 
This is not a restrictive assumption; for any linear function, there is an arbitrarily small perturbation that makes its Hessian diagonalizable; 
in fact, the set of nondiagonalizable matrices over $\mathbb{C}$ has Lebesgue measure~0. 
In comparison, Azizian et al.~\cite{Azizian2020smooth} show exponential convergence of \ref{alg:oga} in linear games with the assumption that the Hessian is invertible. 

\subsection{Smooth and Strongly Concave Payoffs}\label{subsec:strongc-smooth}

A payoff function $\payoffi[ij]$ is said to be $\beta$-smooth, for $\beta > 0$, if for all $\strati, \strati' \in X_i, \strati[j], \strati[j]' \in X_j$,
\begin{align}\label{eq:smoothness}
  \norm{\nabla_{\strati}\payoffi[ij](\strati, \strati[j]) - \nabla_{\strati}\payoffi(\strati', \strati[j])} \leq \beta\norm{\strati - \strati'}; \\
  \norm{\nabla_{\strati}\payoffi[ij](\strati, \strati[j]) - \nabla_{\strati}\payoffi(\strati, \strati[j]')} \leq \beta\norm{\strati[j] - \strati[j]'}\nonumber.
\end{align}

An NZSG is said to have $\beta$-smooth payoffs if each payoff function $\payoffi[ij]$ is $\beta$-smooth for every $[i, j] \in E$.
The game is said to be have $\alpha$-strongly concave payoffs if each $\payoffi$ is $\alpha$-strongly concave in~$\strati$. 
In this section, we show that when players use \ref{alg:ga} and \ref{alg:oga} to update their strategies in a game with payoffs that are $\alpha$-strongly concave and $\beta$-smooth,
their strategies converge to a Nash at an exponential rate. 
Throughout this section, we assume that for each player~$i$, $\payoffi$ is twice continuously differentiable. 
Since each $\payoffi$ is differentiable, it has a unique supergradient, $\nabla_{\strati} \payoffi(\strats)$ at a point $\strats$. 


Before stating our main results, we remark on the existence and uniqueness of Nash.
 Since we consider unconstrained NZSGs, Proposition~\ref{prop:nash} does not apply.
Unlike linear NZSGs in Section~\ref{subsec:linear}, where $\strats = \mathbf 0$ is always a Nash, in general, Nash may not exist when the strategy spaces are not compact.
With $\alpha$-strong concavity, however, we do get uniqueness of Nash when one exists.

\begin{restatable}{lemma}{lemnashunique}\label{lemma:strongc-nash-unique}
In an NZSG with $\alpha$-strongly concave payoffs for $\alpha > 0$, if a Nash equilibrium exists, it is unique.
\end{restatable} 

For applications such as GANs, where strategies are parameters of neural networks, strategy spaces are practically compact, and a Nash equilibrium is guaranteed by Proposition~\ref{prop:nash} to exist.

We now state the main results of this section.

\begin{restatable}{theorem}{propgasmoothstrongc}\label{prop:ga-smooth+strongc}
Consider an unconstrained NZSG with payoffs that are twice continuously differentiable, $\alpha$-strongly concave and $\beta$-smooth for $\alpha, \beta > 0$.
Assume the existence of a Nash, $\strats^*$. Let $\strats^0 \in X$ be such that $\forall i \in V, \norm{\strati^0 - \strati^*} \leq r$ for $r > 0$.
If each player uses \ref{alg:ga}, with $\eta = \frac{\alpha}{2n\beta^2}$, then at each time step~$t$, 
\begin{align*}
    \sum_i\norm{\strati^{t} - \strati^*}^2 \leq  \left(1 - \frac{\alpha^2}{4n\beta^2}\right)^{t}nr^2.
\end{align*}
\end{restatable}

\begin{restatable}{theorem}{propogastrongcsmooth}\label{prop:oga-smooth+strongc}
Consider an unconstrained NZSG with payoff functions that are twice continuously differentiable, $\alpha$-strongly concave and $\beta$-smooth, for $\alpha, \beta > 0$.
Assume the existence of a Nash, $\strats^*$. Let $\strats^0, \strats^1 \in X$ be such that $\forall i \in V, \norm{\strati^0 - \strati^*} \leq r, \norm{\strati^1 - \strats^*} \leq r$ for $r > 0$. If each player uses \ref{alg:oga}, with $\eta = \frac{1}{2n\beta}$, then at each time step~$t$,
\begin{align*}
    \sum_i\norm{\strati^{t+1} - \strati^*}^2 \leq  \left(1 - \frac{\alpha}{4n\beta}\right)^{t}(n+1)2r^2.
\end{align*}
\end{restatable}




In order show convergence for \ref{alg:ga}, we use a Lyapunov-style convergence argument. 
For two-player zero-sum games with strongly-concave and smooth payoffs, Liang and Stokes~\cite{liang2018interaction} show that, when players use \ref{alg:ga} to update their strategies, the strategies converge to the Nash of the game at an exponential rate. 
The key that allows us to extend the result to NZSGs is Lemma~\ref{lemma:util-sum}, which causes terms that are introduced by the strong concavity condition to vanish.

For the \ref{alg:oga} update rule, we make use of writing \ref{alg:oga} as a two step update, so that the second iterate results in a \ref{alg:ga} style update,
\begin{align}\label{alg:oga'}
    \secondstrati^t &= \secondstrati^{t-1} + \eta \nabla_{\strati} \payoffi(\strats^t), \tag{OGA$'$}\\
    \strati^{t+1} &= \secondstrati^t + \eta \nabla_{\strati} \payoffi(\strats^t).\nonumber
\end{align}
Plugging in for $\secondstrati^t, \secondstrati^{t-1}$ in terms of $\strati^t, \strati^{t-1}$ gives us the original \ref{alg:oga} update. 

Mokhtari et al.~\cite{mokhtari2019unified} show that in a two-player zero-sum game with smooth and strongly concave payoffs, if each player uses the  \ref{alg:oga} update, the strategies converge to a Nash exponentially fast. 
Lemma~\ref{lemma:util-sum} again plays a key role in our extension of the result to network zero-sum games.


Azizian et al.~\cite{Azizian2020smooth} show exponential convergence to a Nash when players use the \ref{alg:oga} update strategy in a game with smooth payoffs and ``strongly monotone" dynamics. 
We show in the supplementary file that 
NZSGs with strongly concave payoffs are in fact strongly monotone; this constitutes an alternative derivation of exponential convergence of the OGA dynamics.


\subsection{Lipschitz and Strongly Concave Payoffs}\label{subsec:strongc+lip}

In this section, we show that if players use \ref{alg:ga} or \ref{alg:oga} to update their strategies in an NZSG where payoffs are $\alpha$-strongly concave and $L$-Lipschitz, for $\alpha, L > 0$,
then, given appropriate step sizes, their strategies converge to the unique Nash of the game. 
We assume that for each player~$i$, $\payoffi$ is continuously differentiable. If each $\payoffi$ is $L$-Lipschitz, then for each player~$i$,
\begin{align}\label{eq:Lip-eig}
    \forall \strats \in X, \norm{\nabla_{\strati}\payoffi(\strats)} \leq L.
\end{align}

\begin{restatable}{theorem}{propgastrongclip}\label{prop:ga-lip+strongc}
Consider an unconstrained NZSG that is played for $T$ rounds. Assume each $\payoffi$ is $\alpha$-strongly concave in $X_i$ and $L$-Lipschitz for $\alpha, L > 0$. 
Assume the existence of a Nash, $\strats^*$. 
Let $\strati^0$ be such that, for each player~$i$, $\norm{\strati^0 - \strati^*} \leq r$ for $r > 0$. 
If each player uses \ref{alg:ga} with variable step size $\eta_s > 0$ at each time step~$s$, then at each time step~$t$,
\begin{align*}
    \sum_i\norm{\strati^{t} - \strati^*}^2 \leq L^2n\sum_{s=1}^{t}\eta^2_s + nr^2\prod_{s=1}^{t}\left(1 - \eta_s\alpha \right).
\end{align*}
In particular,  if $\eta_s = T^{-(0.5 + \eps)}$ for $\eps \in (0, 0.5)$, then
\begin{align*}
    \lim_{T \to \infty}\sum_i\norm{\strati^{T} - \strati^*}^2 = 0.
\end{align*}
\end{restatable}

\begin{restatable}{theorem}{propogastrongclip}
Consider an unconstrained NZSG that is played for $T$ rounds. Assume each $\payoffi$ is $\alpha$-strongly concave in $X_i$ and $L$-Lipschitz for $\alpha, L > 0$. 
Assume the existence of a Nash, $\strats^*$. 
Let $\strati^0$ be such that for each player~$i$, $\norm{\strati^0 - \strati^*} \leq r$ for $r > 0$. Then if each player uses \ref{alg:oga} with nonincreasing step size $\eta_s > 0$,
\begin{align*}
    \sum_i\norm{\strati^{t} - \strati^*}^2 \leq 4 nL^2\sum_{s=1}^{t}\eta_s\eta_{s-1} + nr^2\prod_{s=1}^{t}(1 - (\eta_s + \eta_{s-1}) \alpha).
\end{align*}
In particular,  if $\eta_s = T^{-(0.5 + \eps)}$ for $\eps \in (0, 0.5)$, then
\begin{align*}
     \lim_{T \to \infty}\sum_i\norm{\strati^{T} - \strati^*}^2 = 0.
\end{align*}
\end{restatable}




Our proofs for these theorems resemble those from Section~\ref{subsec:strongc-smooth}, with Lemma~\ref{lemma:util-sum} facilitating the generalization to NZSGs. 
We note that the proof fails to achieve exponential convergence, due to the lack of smoothness in the game. 
Furthermore, the algorithm designer needs to know in advance the time horizon~$T$, the number of time steps the game is to be played, in order to choose a learning schedule that allows for guaranteed last-iterate        convergence.

\section{Experiments}
\label{sec:exp}
In this section, we provide examples validating our results. We first show convergence in the simplest setting --- a game with three players where a zero-sum game is played between each pair. We provide experiments showing convergence in a game with linear payoffs, and a game with smooth and strongly concave payoffs. We then provide an experiment showing the effect that increasing the number of players has on convergence rate. For each experiment, we show the performance of both \ref{alg:ga} and \ref{alg:oga}.
\subsection{Three Player Game with Linear Payoffs}

\begin{figure}[H]
\centering
   \begin{subfigure}{0.4\linewidth} \centering
     \includegraphics[scale=0.3]{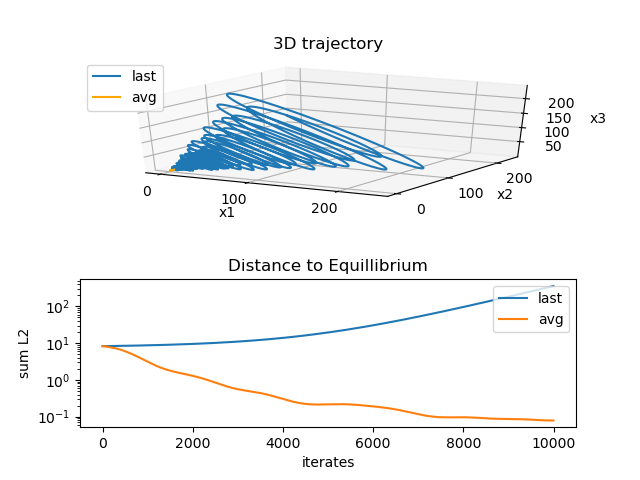}
     \caption{\ref{alg:ga}}\label{fig:figA}
   \end{subfigure}
   \begin{subfigure}{0.4\linewidth} \centering
     \includegraphics[scale=0.3]{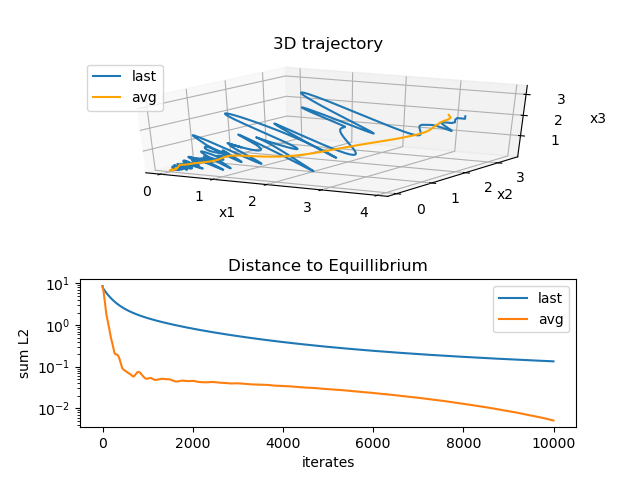}
     \caption{\ref{alg:oga}}\label{fig:figB}
   \end{subfigure}
\caption{In the linear game, \ref{alg:ga} last iterate diverges, while \ref{alg:ga} average iterate converges. For \ref{alg:oga}, both last and average iterate converge.} \label{fig:bilinear}
\end{figure}

We provide experiments validating our theoretical results for a three player game with linear payoffs. The payoff of the players can be expressed as
\begin{align*}
    \begin{bmatrix} 
    \payoffi[1](\strats) \\ \payoffi[2](\strats) \\
    \payoffi[3](\strats) 
    \end{bmatrix} = \begin{bmatrix} \strati[1]^{\top} & \strati[2]^{\top} & \strati[3]^{\top} \end{bmatrix} \begin{bmatrix} 0 &  C_{12} & C_{13} \\
    -C_{12}^{\top} & 0 & C_{23} \\
    -C_{13}^{\top} & - C_{23}^{\top} & 0
    \end{bmatrix} \begin{bmatrix} \strati[1] \\ \strati[2] \\ \strati[3] \end{bmatrix}.
\end{align*}

To track convergence, it is convenient if the game has a unique Nash. The linear game will have a unique Nash at $\mathbf{0}$ as long as the fixed point of dynamics is the singleton $\{\mathbf{0}\}$. This will occur if the Hessian of payoffs, $H$, has no eigenvalues equal to 0. Since $H$ is antisymmetric, its eigenvalues come in complex pairs, and if $d$ is even dimension, $H$ will have an eigenvalue equal to 0 if its determinant is 0. If we sample entries of the $C_i$'s i.i.d from the uniform distribution this will happen with probability 0.
 
We let $C_{ij} \in \mathbb{R}^{10 \times 10}$, and initialize the entries by sampling i.i.d. from the uniform distribution on $[0,1]$.
We initialize the coordinates of $\strati[1]^0, \strati[2]^0, \strati[3]^0$ i.i.d from the uniform distribution on $[-1,1]$. For \ref{alg:ga} we set $\eta=0.003$, to allow us to visualize the convergence of the average iterate and the divergence of the last iterate on the same plot. For \ref{alg:oga} we let $\eta=0.05$ for fastest convergence. We plot the trajectory of a single representative game simulation. 

We demonstrate the performance of \ref{alg:ga} and \ref{alg:oga} by plotting the trajectories of players' strategies; to show this in~$\mathbb R^3$, we take the $\ell_2$-norm of each player's strategies to form a three dimensional vector. 
We also plot the the sum of the squares of the $\ell_2$ distance of player strategies from the origin on a log scale. This is shown in Figure~\ref{fig:bilinear}. From our results, it can be seen that \ref{alg:ga} diverges from the unique Nash of the game, while \ref{alg:oga} converges to the unique Nash in last iterate. Notice that although the last iterate of \ref{alg:oga} converges, it does so at a slower speed than the average iterate. The convergence in last iterate of \ref{alg:oga} is not quite linear, but is upper bounded by a linear function, and hence does not contradict our theory.

\subsection{Three Player Game with Smooth and Strongly Concave Payoffs}\label{subsec:exp-strongc}

\begin{figure}[H]
\centering
   \begin{subfigure}{0.49\linewidth} \centering
     \includegraphics[scale=0.3]{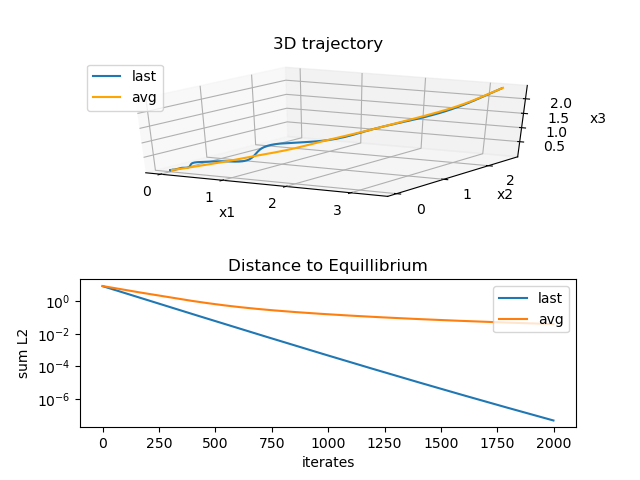}
     \caption{\ref{alg:ga}}\label{fig:figC}
   \end{subfigure}
   \begin{subfigure}{0.49\linewidth} \centering
     \includegraphics[scale=0.3]{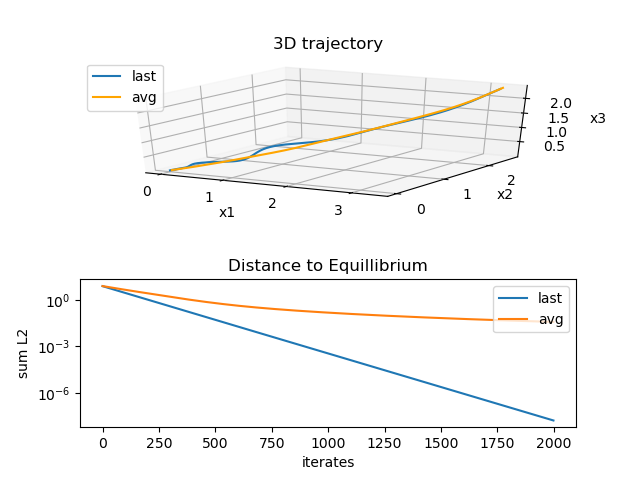}
     \caption{\ref{alg:oga}}\label{fig:figD}
   \end{subfigure}
\caption{In the smooth and strongly concave game, last iterate and average iterate converges for both algorithms.} \label{fig:smooth}
\end{figure}

We next provide experiments showing convergence in a three player game with smooth and strongly concave payoffs. We set the payoffs for each player as follows:
\begin{align}\label{eq:smooth+strongc}
    &\payoffi[ij](\strats) = -\frac{1}{2}\norm{\strati}^2 + \strati^{\top}C_{ij}\strati[j] + \frac{1}{2}\norm{\strati[j]}^2\\
    &\payoffi(\strats) = \sum_{j \in V\backslash\{i\}}\payoffi[ij](\strats)\notag
\end{align}
Like in the game with linear payoffs, this game has a unique Nash at $\mathbf{0}$ if and only if the determinant of $H$ is nonzero, which we can guarantee by sampling entries of the $C_i$'s uniformly at random.

As in the linear game, we initialize the entries of $C_{ij} \in \mathbb{R}^{10 \times 10}$
by sampling i.i.d.\@ from the uniform distribution on $[0,1]$, and the coordinates of $\strati[1]^0, \strati[2]^0, \strati[3]^0$ i.i.d from the uniform distribution on $[-1,1]$. We set $\eta=0.005$ for both \ref{alg:ga} and \ref{alg:oga}. The results are shown in Figure~\ref{fig:smooth}. 

From our plots, it can be seen that both \ref{alg:ga} and \ref{alg:oga} last iterates converge for the smooth and strongly concave game. 
Although \ref{alg:oga} converges for both the game with linear payoffs and the game with smooth and strongly concave payoffs, the trajectory of \ref{alg:ga} and \ref{alg:oga} take a more direct path to the Nash in the smooth and strongly concave game, as can be seen in the 3d trajectory. Furthermore, the last iterate of both \ref{alg:ga} and \ref{alg:oga} follow a linear trend in the log scale, as predicted by our theory.

\subsection{Effect of Number of Players on Convergence}

\begin{figure}[H]
\centering
\includegraphics[width=0.7\textwidth]{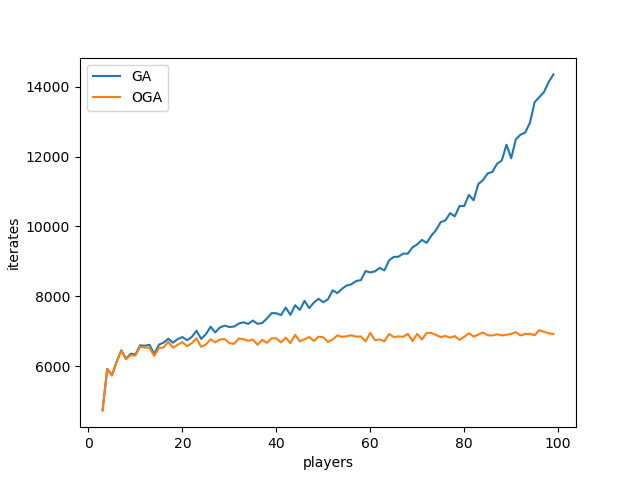}
\caption{In the smooth and strongly concave game, the performance of \ref{alg:ga} decays as players increase, while the performance of \ref{alg:oga} plateaus.} \label{fig:players}
\end{figure}

In this section, we provide experimental results showing the effect of varying $n$ in a NZSG of $n$ players. For the smooth and strongly concave game, our theoretical upper bound has a linear dependence on the number of players in the game for both \ref{alg:ga} and \ref{alg:oga}, and thus we test the dependence on players only in this setting.

We study a game with smooth and strongly concave payoffs, using the same payoffs as in Section~\ref{subsec:exp-strongc} (see Equation~\eqref{eq:smooth+strongc}). We perform the same initializations as in Section~\ref{subsec:exp-strongc} - initializing the entries of $C_{ij} \in \mathbb{R}^{10 \times 10}$
by sampling i.i.d.\@ from the uniform distribution on $[0,1]$, and the coordinates of $\strati^0$ i.i.d from the uniform distribution on $[-1,1]$. We set $\eta=0.001$ for both \ref{alg:ga} and \ref{alg:oga}. We let the number of players range from 3 to 100, plotting convergence for each setting of players. We track convergence by plotting the number of iterates it takes for the the sum of the squares of the $\ell_2$ distance of player strategies from the origin to dip below $0.00001$. For each fixed number of players, we run ten trials to convergence and plot the average. The results are shown in Figure~\ref{fig:players}. 

From this plot, we can see that the number of players affects the convergence rate of \ref{alg:ga}. However, for \ref{alg:oga}, the effect of players on convergence disappears after enough players are introduced into the game. This suggests that the convergence rate for \ref{alg:oga} in the smooth and strongly concave case may not be tight. This is an open question for future research.

\section{Conclusion}

 In this paper, we studied the convergence of player strategies to equilibria in Network Zero-sum Games, a class of games that generalizes two-player zero-sum games and arises naturally in learning architectures that extend GANs.
 We show that many results in two-player zero-sum games on the convergence and divergence of these algorithms extend to NZSGs.
 We believe these results may guide practitioners working on extensions of GANs that involve more than two agents.
 Our results also shed some light on why existing extensions of GANs that employ more than two agents are successful in achieving convergent behaviour. 
 Future research may search for models with more relaxed game theoretic assumptions where convergence can still be shown for reasonable algorithms. 
 For example, the zero-sum assumption 
 is absent from certain successful architectures, e.g.\@ Wasserstein-GAN with Gradient Penalty \cite{wei2018improving}.


\end{document}